\documentclass[letterpaper, 10 pt, conference]{ieeeconf}  
\usepackage[table,xcdraw]{xcolor}
\usepackage[normalem]{ulem}
\useunder{\uline}{\ul}{}
\usepackage{xcolor}
\usepackage{graphicx}
\usepackage{amsmath}
\usepackage{algorithm}
\usepackage{dblfloatfix} 
\usepackage[noend]{algpseudocode}
\usepackage[utf8]{inputenc}
\usepackage{bm}
\algdef{SE}[SUBALG]{Indent}{EndIndent}{}{\algorithmicend\ }%
\algtext*{Indent}
\algtext*{EndIndent}
\algdef{SE}[FOR]{NoDoFor}{EndFor}[1]{\algorithmicfor\ #1}{\algorithmicend\ \algorithmicfor}%
\algdef{SE}[IF]{NoThenIf}{EndIf}[1]{\algorithmicif\ #1}{\algorithmicend\ \algorithmicif}%
\IEEEoverridecommandlockouts                              
                                                          
\title{\LARGE \bf	
Scene Retrieval for Contextual Visual Mapping
}

\author{William H. B. Smith $^{1}$, Michael Milford $^{2}$, Klaus D. McDonald-Maier $^{1}$ and Shoaib Ehsan $^{1}$
\thanks{This work was supported by the UK Engineering and Physical Sciences Research Council through Grants EP/R02572X/1, EP/P017487/1 and in part by the RICE project funded by the National Centre for Nuclear Robotics Flexible Partnership Fund.}
\thanks{$^{1}$W. H. B. Smith, K. McDonald-Maier and S. Ehsan are with the School of Computer Science and Electronic Engineering, University of Essex, Colchester CO4 3SQ, United Kingdom. (e-mail: ws20956@essex.ac.uk; kdm@essex.ac.uk; sehsan@essex.ac.uk)}%
\thanks{ $^{2}$ M. Milford is with the School of Electrical Engineering and Computer Science, Queensland University of Technology, Brisbane, QLD 4000, Australia. (e-mail: michael.milford@qut.edu.au)
        }%
}

\begin{document}

\maketitle
\thispagestyle{empty}
\pagestyle{empty}

\begin{abstract}
Visual navigation localizes a query place image against a reference database of place images, also known as a `visual map'. Localization accuracy requirements for specific areas of the visual map, `scene classes', vary according to the context of the environment and task. State-of-the-art visual mapping is unable to reflect these requirements by explicitly targetting scene classes for inclusion in the map. Four different scene classes, including pedestrian crossings and stations, are identified in each of the Nordland and St. Lucia datasets. Instead of re-training separate scene classifiers which struggle with these overlapping scene classes we make our first contribution: defining the problem of `scene retrieval'. Scene retrieval extends image retrieval to classification of scenes defined at test time by associating a single query image to reference images of scene classes. Our second contribution is a triplet-trained convolutional neural network (CNN) to address this problem which increases scene classification accuracy by up to 7\% against state-of-the-art networks pre-trained for scene recognition. The second contribution is an algorithm `DMC' that combines our scene classification with distance and memorability for visual mapping. Our analysis shows that DMC includes 64\% more images of our chosen scene classes in a visual map than just using distance interval mapping. State-of-the-art visual place descriptors AMOS-Net, Hybrid-Net and NetVLAD are finally used to show that DMC improves scene class localization accuracy by a mean of 3\% and localization accuracy of the remaining map images by a mean of 10\% across both datasets. 
\end{abstract}

\section{INTRODUCTION}





Visual navigation localizes a query place image against a reference set of place images, referred to in this paper as a `visual map'. Current approaches add images to this map according to: time \cite{Warren} and distance \cite{Sourav2017} intervals, distinctiveness \cite{Chapoulie}, or a tri-folded criteria \cite{Zaffar2020A} which we will refer to as `memorability’. Of these approaches only memorability includes images in the visual map according to any explicit, external image criteria. However, memorability does not account for the localization requirements of the `scene class' represented in that image. 

\begin{figure}[h!]
	\centering
	\includegraphics[scale=0.165]{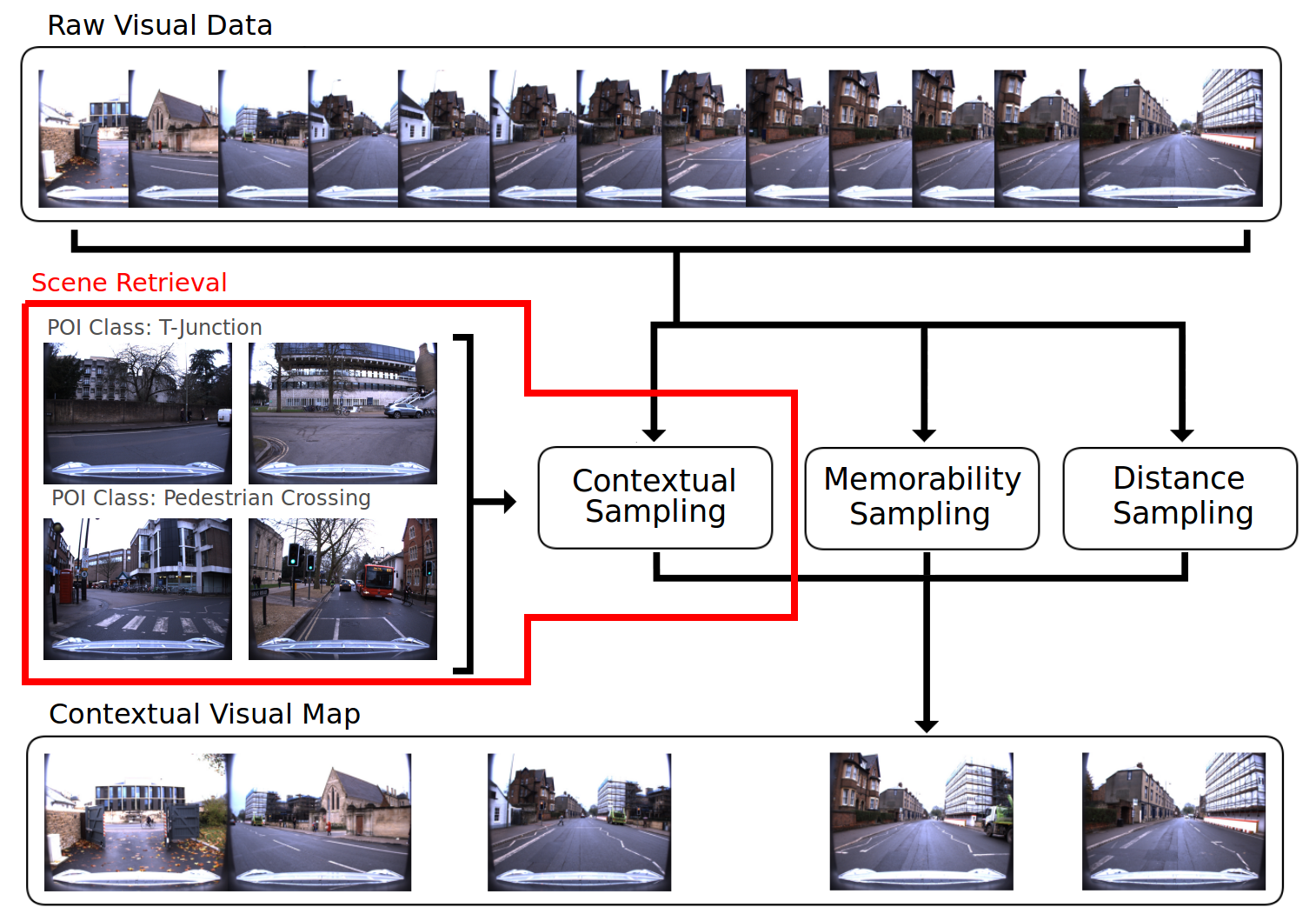}
	\caption{DMC combines distance, memorability and context from scene retrieval for contextual visual mapping to include more user defined classes in visual maps}
	\label{fig:CSC_system}
\end{figure}

In contrast our approach allows a user to increase localization accuracy requirements for target scene classes by providing examples of them to classify incoming images against. The resulting visual map includes more instances of the target scene classes compared to selecting them using memorability or distance intervals. The improved visual maps then allow more accurate localization of those scene classes.



\begin{figure}[h!]
	\centering
	\includegraphics[scale=0.38]{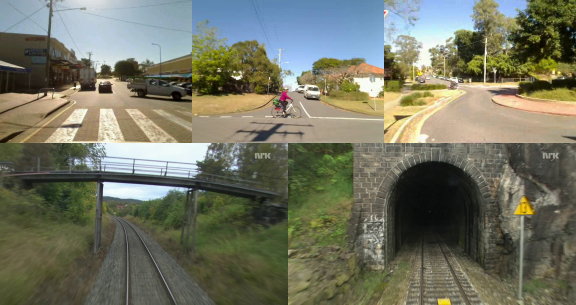}
	\caption{Scene class examples. Top three images are from St. Lucia. Left to right: pedestrian crossing, T-junction and roundabout. Bottom two images are from Nordland. Left to right: bridge and tunnel.}
	\label{fig:class samples}
\end{figure}

Scene classes (Figure \ref{fig:class samples}) are selected by the user according to the environment and task at hand. Four scene classes for three traversals of the St. Lucia \cite{Glover2010ICRA} and Nordland \cite{Sunderhauf2013} datasets were selected. St Lucia is a 18km route of an Australian suburb at recorded at different times of day and with scene classes of: pedestrian crossings, roundabouts, four-way junctions, and T-junctions. Nordland is a 729km rail journey through rural and suburban Norway. Bridges, level crossings, stations and tunnels were the scene classes selected from spring, summer and autumn traversals. All remaining images in both datasets were assigned to a fifth scene class `undefined'. Traditionally scene recognition would teach a classifier on these classes, but that has two notable weaknesses. Firstly, it would need to be re-trained for each environment and set of scene classes and secondly the visual overlap between scenes such as T-junctions and pedestrian crossings makes classifier training very challenging. 

In response this paper takes inspiration from image retrieval to formulate the `scene retrieval' problem: classification of a query image by comparing it to unlimited examples of different user-defined scene classes (Figure \ref{fig:CSC_POI_detect}) at test time. The scene class with the average smallest distance (as defined below) between its example images and the test image descriptors becomes the classification. This problem formulation aims to eliminate the need for re-training a descriptor on new classes or environments. Our solution proposes a convolutional neural network (CNN) trained using triplet loss to generate embedded image descriptors for scene retrieval. Triplet learning is used to leverage the visual similarity between classes to improve training. Our approach increases classification accuracy by up to 7\% in comparison to descriptors from state-of-the-art scene recognition networks \cite{Zhou2014} pre-trained on Places365 \cite{Zhou2017} and ImageNet datasets.

Furthermore, the contextual information provided by this paper's solution to scene retrieval is combined with distance and memorability \cite{Zaffar2020A} in our visual mapping algorithm: `DMC' (Figure \ref{fig:CSC_system}). DMC creates visual maps which include up to 64\% more images of the specified scene classes than distance mapping. The inclusion of memorability in DMC is also shown to improve scene class localization accuracy by a mean of 3\% and localization accuracy of the remaining map images by a mean of 10\% across both datasets using state-of the art visual place recognition (VPR) features AMOS-Net, Hybrid-Net \cite{Sunderhauf2017} and NetVLAD \cite{Arandjelovi}. This paper's contributions are:

\begin{enumerate} 
	\item Formulation of a 'scene retrieval' problem to classify overlapping scenes defined at test time.
	\item A solution to scene retrieval that uses a triplet trained CNN that increases scene classification accuracy by up to 7\% in comparison to the state-of-the-art scene recognition models.
	\item A visual mapping algorithm, `DMC' that uses context from scene retrieval, memorability and distance to include 64\% more instances of user-defined scene classes in a visual map, in comparison to distance interval mapping. Our analysis shows the inclusion of memorability also increases scene class localization accuracy by a mean of 3\% and localization accuracy of the remaining map images by a mean of 10\% in comparison to just using context.
\end{enumerate}

\section{BACKGROUND}

\subsection{Scene Recognition}
Our approach is most closely related to hybrid deep models for scene recognition \cite{Xie}. DAG-CNN \cite{Songfan} integrates features from different levels of a CNN in a directed acrylic graph and FOSNet \cite{Seong} introduces scene coherence loss to fuse object and scene data for improved performance. CNN architectures used for object recognition have also been adapted and re-trained for place recognition \cite{Zhou2017}. These architectures were used for comparison against our scene retrieval models.

Some scene classes described in this paper overlap with autonomous driving events \cite{Yurtsever}, such as stopping at a junction. However, events usually detected such as lane changing, overtaking and rear-ending \cite{Xue} cannot be directly linked with the scene classes in this paper.

\subsection{Visual Mapping}
Visual navigation data is typically collected using videos at a constant frame rate. Therefore a time interval is the most convenient way to create visual maps \cite{Warren}. Datasets that include position data can be mapped at distance intervals \cite{Sourav2017}, but are dependent on the accuracy of the position data which is typically provided by GPS. Images can also be mapped according to distinctiveness \cite{Chapoulie}, but user-defined emphasis on specific types of distinctiveness is not possible.

Zaffar et al. \cite{Zaffar2020A} consolidate metrics such as feature stability \cite{Dymczyk} and matchability \cite{Hartmann} to predict the suitability of images for visual mapping by selecting visual map images based on their memorability, staticity and entropy. Entropy is calculated on a per-pixel basis, staticity is measured by using YOLOv2 \cite{Redmon} to detect occlusions from objects such as pedestrians or cars and a second network pre-trained on human memorability \cite{Khosla} is used to predict memorability.  

Context for VPR is restricted to local visual cues detected using attention models \cite{Khaliq2019A} \cite{Chen} or by concatenating features from different CNN layers \cite{Chen2017A}. Object detection is also used to deduce visual context in new environments \cite{Pal}.

\subsection{Visual Place Recognition}
Visual place recognition (VPR) is framed as image retrieval where representations of reference example images (visual map) are compared with test image representations \cite{Mubariz2019A}. Many descriptors have been designed for this task to be invariant to viewpoint \cite{Khaliq2019} \cite{Mubariz2019B} \cite{Garg2019A}, environmental \cite{Sunderhauf2017} \cite{Zaffar2020} and long-term visual \cite{Wang2019} changes. 

Triplet learning \cite{Hoffer} is used to leverage the overlap between place images for learning descriptors \cite{Lopez-Antequera}. Variations on triplet learning have also been used to improve training data selection and select the most relevant features for VPR \cite{Hausler}. We extend the use of triplet learning to learn the relationship between classes of images for scene recognition, rather than single images in different conditions for VPR. 

Additionally this paper uses three state-of-the-art VPR descriptors: AMOS-Net, Hybrid-Net \cite{Sunderhauf2017} and NetVLAD \cite{Arandjelovi} to show the positive effect of combining memorability with contextual and distance interval mapping for localization. AMOS-Net is a modified version of Caffe-Net with all parameters trained for VPR, whereas Hybrid-Net's top 5 convolutional layers were initialized from Caffe-Net. `Conv5’ layer features from both networks are extracted and encoded using Spatial Pyramidal Pooling. NetVLAD appends a VLAD layer to a partially frozen VGG16 network pre-trained on the ImageNet dataset retrained for VPR using the aforementioned triplet learning. NetVLAD features are reduced to 128 values using PCA. 



\section{METHOD}
This section describes the scene classes, the problem formulation of scene retrieval and how our neural network model was used to address it. It then describes the three visual mapping approaches used for scene classification and our combination of all three in our DMC algorithm.

\subsection{Scene Classes}

The scene classes were chosen for each dataset based on their repeatability and general applicability for navigation, with a focus on safety. For St. Lucia the scene classes were: pedestrian crossings, roundabouts, four-way junctions, and T-junctions. For Nordland the scene classes were bridges, level crossings, stations and tunnels. For tunnels both the approach and exit from the tunnel was used, when there was a sufficient view of the outside. Bridge classes in Nordland were from passing under bridges. Scene classes were hand labelled in both datasets

\subsection{Scene Retrieval: Problem Formulation}
\label{SceneRetBegin}
Classification of scenes defined at test time is needed to provide context for task-specific visual mapping, where a visual map is defined as a collection of images used to represent a geographical area. Two current approaches could be used for scene classification.

\subsubsection{Scene Recognition}
Scene recognition is classification of scenes defined at training time. A classifier could be trained on our scene classes. However, the visual overlap between classes from a single visual map made this very challenging. It would also need to be re-trained if the scene classes or visual map environment were to change.

\subsubsection{Visual Place Recognition}
VPR is defined as image retrieval. A descriptor is used to represent a query image which can then be classified by comparing it to an unlimited amount of individual user-defined classes (i.e. scenes) represented using the same type of descriptor. 

\subsubsection{Scene Retrieval}
Scene retrieval is defined as: classification of overlapping scene classes defined at test time. This also requires a solution that does not need re-training for each class or new environment and is different from VPR because it is designed to match images of one scene class (i.e. roundabout) and multiple images of different scene classes (i.e. roundabouts, bus stops, etc.) rather than to match two individual scene images in different visual conditions.

\begin{figure}[h!]
	\centering
	\includegraphics[scale=0.165]{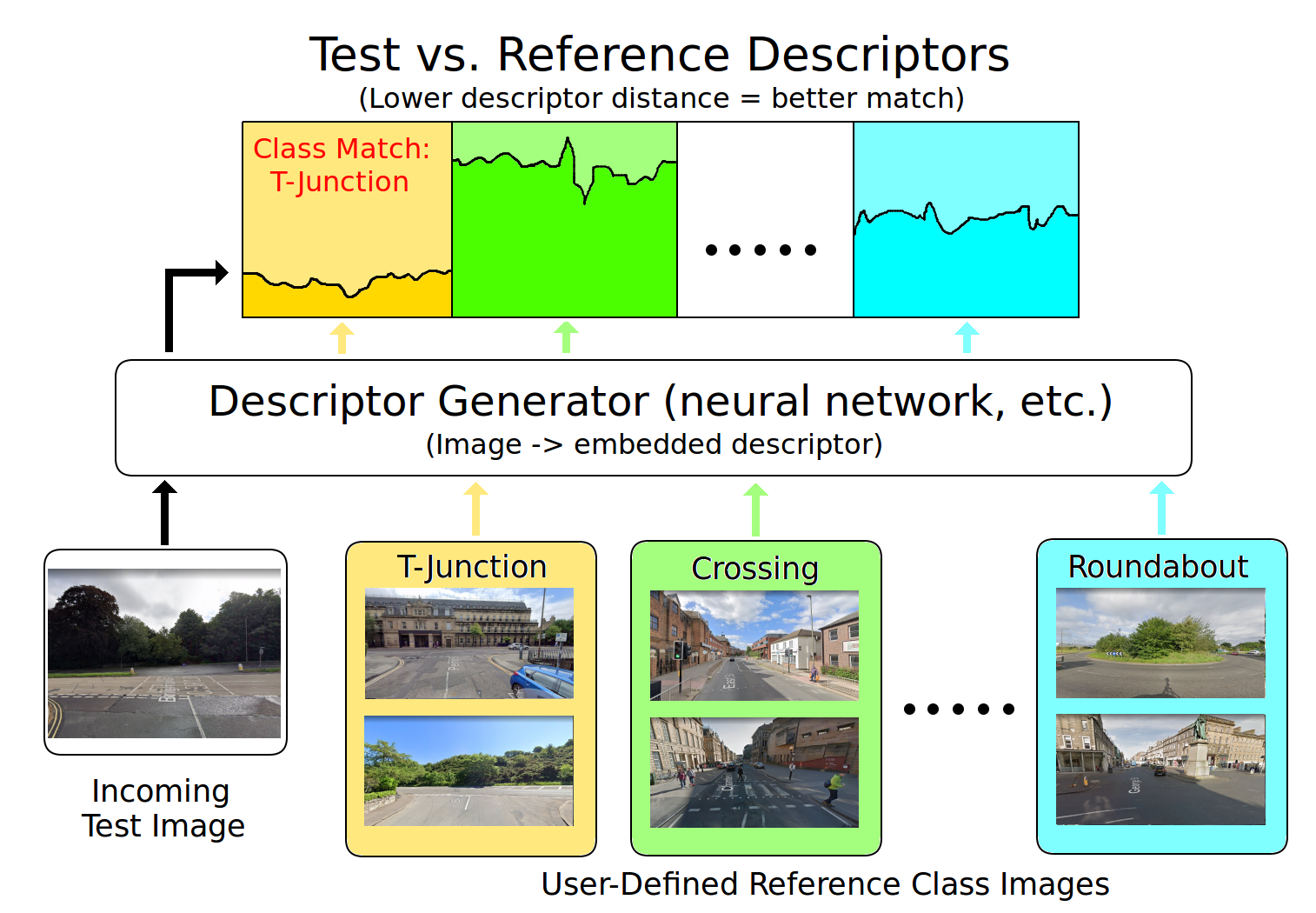}
	\caption{Scene retrieval pipeline. The reference class scene whose image descriptor is the average, shortest Euclidean distance from the query image's descriptor becomes the predicted scene.}
	\label{fig:CSC_POI_detect}
\end{figure}

This paper's solution uses triplet loss to train a neural network to generate embedded descriptors for scene classes. At test time the reference scene class whose embedded descriptors are, on average, the shortest Euclidean distance from the query image's descriptor is used as the classification result (Figure \ref{fig:CSC_POI_detect}).

Scene retrieval is a challenging problem that is only introduced in this paper so we confine ourselves to taking reference images of each scene class from one dataset and comparing them against the remaining, previously unseen, scene images from the same dataset. This simulates a user hand labelling a limited subset of scene classes in a dataset and then using them to identify the remaining examples of those scene classes. Scenes from other datasets could be used for reference, but we leave that for future work.

\subsection{Scene Retrieval: Training a Model}
\label{ScenRetEnd}

\subsubsection{Triplet Learning and Architecture}
State-of-the-art place recognition approaches used for comparison on this task \cite{Zhou2017} make use of a VGG16 architecture. For scene retrieval we use a triplet network formed by using three instances of the VGG16 architecture with the final classification layer removed and shared weights. 

\begin{figure}[h!]
	\centering
	\includegraphics[scale=0.26]{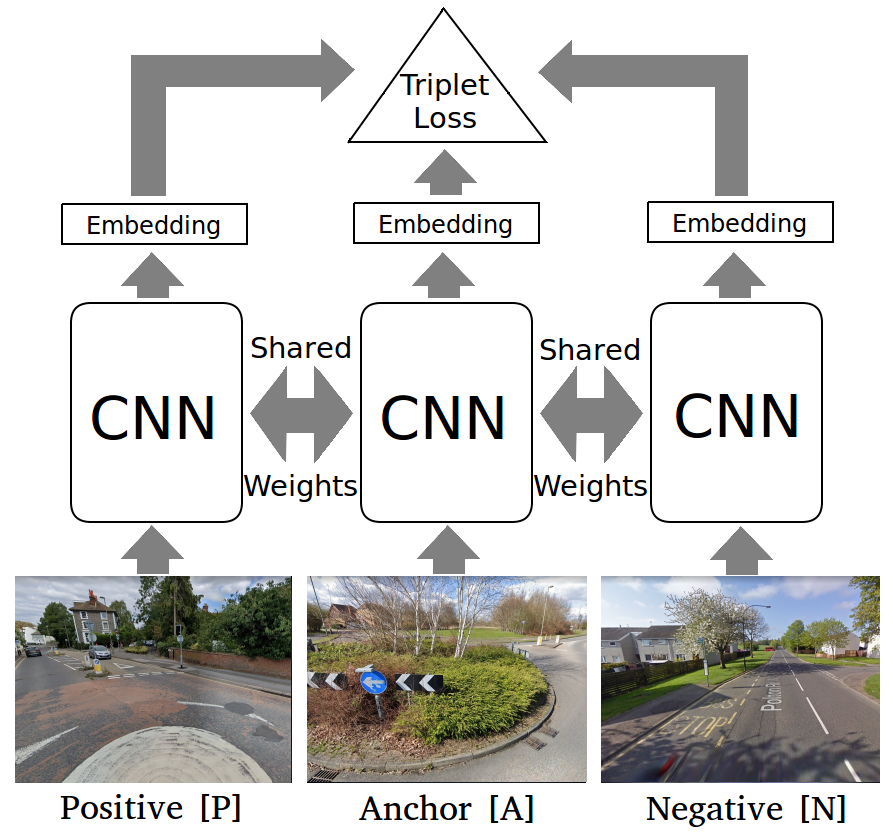}
	\caption{Example of a triplet mined from the EE Road Dataset and network architecture diagram. Positive and anchor images are of roundabouts, the negative image is of a bus stop.}
	\label{fig:triplet}
\end{figure}

We train the network to produce accurate representations of the scenes by passing three images through the network: a randomly chosen `anchor' image, a different `positive' image from the same scene class and a randomly chosen `negative' image from a different scene class. Each image is passed through the separate instances of the network to create three embedded representations. Triplet loss is then used to  minimize the Euclidean distance between the anchor and positive and maximize it between the anchor and negative. A visual example of this process is shown in Figure \ref{fig:triplet}.

\subsubsection{Training Data}
\label{training}
A 700 image dataset (EE Road Dataset) was collected by hand from the Scottish capital city Edinburgh and London satellite town Epsom using Google Street View. The dataset contains 100 GPS-tagged examples of 5 semi-urban scene classes: pedestrian crossing, roundabout, bus stop, four-way junction and T-junction. A 6th class of 200 images containing none of those classes is also included. A 10km route from the Oxford RobotCar dataset \cite{Maddern} was also hand labelled for the same scene classes, minus 'roundabout', and used as training data. The Places 2 \cite{Zhou2014} dataset contains approximately 8 million examples of 365 scene classes, but the intra-class relationship was significantly different from the labelled Oxford RobotCar and EE Road datasets. A 365,000 image subset of Places 2 (1000 from each class) was used for training, but not combined with the two previous datasets as this did not increase performance. 

Positive results were achieved with a small training set based on combinations of VPR data. Similar success on the same test set shown in Section \ref{SceneRet} was found by adapting an existing place recognition dataset. This shows that the increasingly comprehensive VPR, place recognition and object recognition datasets could easily be used to address classification of overlapping classes in a number of different domains, users need only ensure a roughly consistent intra-class relationship between datasets.

\subsubsection{Trained Models}
\label{trained_models}
Our two proposed networks (EEOR, P2) were initialized with the weights from the network in \cite{Zhou2017}, which was originally trained on Places365, and then trained using a triplet margin of 0.2. EEOR was trained on a 50/50 mix of 3.5 million triplets from the Oxford RobotCar and EE Road datasets in batches of 8 spread over 120 epochs. P2 was trained using 2.9 million  triplets mined from the aforementioned subset of the Places 2 dataset in batches of 8 spread over 100 epochs. Training was done on a Nvidia GTX1070 and took 7.5 and 6 hours respectively.

\subsection{Visual Mapping: Distance, Memorability and Context}
\label{sampling_techniques}

\subsubsection{Distance} 
Adding images to the visual map at set distance intervals is a significant improvement over time interval mapping for mapping because it ensures places are represented regardless of the speed they are passed through. It was used as the baseline visual mapping technique for identifying scene classes.

\subsubsection{Memorability}
Memorable Maps \cite{Zaffar2020A} uses three metrics to decide on an images's suitability for inclusion in a visual map. Memorability uses a convolutional neural network taught on place recognition and fine tuned for human memorability, however this may highlight dynamic objects as memorable, despite them not being re-observable. Staticity uses a different convolutional neural network trained on object detection to find and mask dynamic objects. Painted walls are often observed during outdoor navigation and may be classified as both static and memorable, however they typically have no unique features and therefore may lead to false positive matches with other similar walls. Local pixel entropy is used to measure how much information is encoded in each part of the image, allowing images that would lead to false positive matches to be discarded. Memorability is used as the second baseline for selecting scene classes. An image's memorability score is between 1 and 0, the higher the better. For mapping an image's memorability was calculated and if it was above a defined threshold it was added to the map. 


\begin{algorithm}[h!]
	\caption{Distance, Memorability and Contextual Visual Map Sampling}
	\label{alg:my_alg}
	\renewcommand\algorithmicthen{}
	\begin{algorithmic}[1]
		
		\NoDoFor each user-defined scene class, $S_C$, select example images from the map data.
		\EndFor
		
		\NoDoFor each $S_C$ calculate the embedded representation of all its images, $S_{CE}$ using the trained scene retrieval network.
		\EndFor
		
		\State Initialise empty visual map, $V_{map}$.
		
		\State Set memorability bias, $B_{mem}$.
		
		\State Set matching confidence threshold, $threshold_{S}$.
		
		\State Set min. and max. distance allowed between map frames $= dist_{min}$, $dist_{max}$. 
		
		\NoDoFor each remaining image of the map data, $Img$:
		
		\State Calculate the distance moved, $Dist_m$, since the last image was added to $V_{map}$.
		
		\State Calculate the memorability of the image $Img_{mem}$
		
		\State Calculate the scene classification $Img_{S}$ and the confidence score from that class, $Img_{score}$ 
		
		\State $Img_{score} += (Img_{score} * B_{mem} * Img_{mem})$
		
		\State \textbf{if} $Img_{S}$ == `undefined':
		\State \indent  Discard $Img$.
		
		\State \textbf{elif} $Img_{score} < threshold_{S}$: 
		\State \indent add $Img$ to $V_{map}$.
		
		\State \textbf{elif} $Dist_m > dist_{min}$ and $Img_{mem} > threshold_{mem}$:
		\State \indent Add $Img$ to $V_{map}$.
		
		\State \textbf{elif}  $Dist_m > dist_{max}$:
		\State \indent Add $Img$ to $V_{map}$.
		
		\State {\textbf{else:}}
		\State \indent  Discard $Img$.
		
		\EndFor
		
	\end{algorithmic}
\end{algorithm}

\subsubsection{Context}
\label{Context_sampling}
This paper uses scene retrieval to provide context for visual mapping by classifying user-defined scene classes. Images of each scene class are taken from one traversal of one dataset and then compared against the remaining, previously unseen, scene images from the same traversal. When localizing test images against the sampled visual map localization accuracy is not measured for the user-defined scene images which are also removed from the map entirely. 

The output of the scene retrieval pipeline is a scene classification and the average distance between the test image descriptor and that reference scene image descriptors, this is called the confidence score. For contextual mapping, if a frame's confidence score was above a defined threshold and it was not classified as an `undefined' class image it was added to the visual map. 


\subsection{Visual Mapping: Combining Distance, Memorability and Context, 'DMC'} 
Contextual mapping alone creates a visual map of only scene images so it was combined with distance and memorability (Algorithm \ref{alg:my_alg}) to ensure the map covers a wide area and consists of suitable images for localization at undefined scene classes. A query image's contextual confidence score is biased by its memorability, if it is below a threshold then it is added to the map. Otherwise, images are added if they are above the memorability threshold and a minimum distance has been moved, or if the distance moved exceeds a maximum threshold. 


\section{EXPERIMENTS}

\subsection{Scene Retrieval: How Accurate is Scene Classification?}
\label{SceneRet}
The purpose of this experiment was to test the accuracy of our scene classification against the state-of-the-art scene recognition approaches.

The mean number of images per scene class, per example traversal for St Lucia used for this experiment were: pedestrian crossings (28), roundabouts (59), four-way junctions (64), T-junctions (100) and a sample of 100 `undefined' images. For Nordland the scene class counts were: bridges (491), level crossings (431), stations (870), tunnels (531) and a random sample of 2000 `undefined' images. 

A variant of 4-fold validation was used for testing. For each of the two dataset's three traversals the scenes were split into 4 equal, sequential sections and used in turn as reference data while the remaining images from the same traversal were used as test data and compared using the trained models `EEOR' and `P2' described in Section \ref{trained_models} and pipeline illustrated in Figure \ref{fig:CSC_POI_detect}. 

The same experiment was then repeated using two state-of-the-art networks `365' and`1365' \cite{Zhou2017} trained for place recognition. As neither of these networks natively support classification of the scene classes defined by this paper both had their final classification layer removed and were then substituted for the `descriptor generator' in Figure \ref{fig:CSC_POI_detect}. '365' was taught only on Places365 data, while '1365' was also taught on the ImageNet dataset. The classification results were averaged and summarized in Tables \ref{table:POI_detection_STL} \& \ref{table:POI_detection_N}.


\subsubsection{How Well Do Our Models Generalize?}
\label{GPSR}
EEOR outperforms the state-of-the-art for classifying scenes that it was trained to recognize on a different dataset and P2 outperforms the state of the art on both datasets for entirely unknown scenes without re-training. P2 shares the same number of parameters as '365' and was taught on a subset of its original training data so proves the efficacy of our approach as it outperforms `365', by an average of 6\% higher classification accuracy across both datasets, 

Classification on Nordland is more successful than St. Lucia possibly because the St. Lucia dataset only covers a relatively uniform suburban area so its scene class overlap is higher.
\begin{table}[h!]
	\caption{4 Class (see Section \ref{open}) scene classification success (\%) on the St. Lucia dataset: state-of-the-art vs. ours.}
	\centering
	\label{table:POI_detection_STL}
	\scalebox{1.2}{
		\begin{tabular}{l|l|l|l|l}
			& \hspace{.0cm} \textbf{365} \hspace{.0cm} & \textbf{1365} & \hspace{.1cm}\textbf{P2}& \textbf{EEOR} \\\hline
			\textbf{Ped. Crossing} & \hspace{.15cm}65  & \hspace{.15cm}\underline{67}   & \hspace{.12cm}58                & \hspace{.25cm}59                \\
			\textbf{Roundabout} & \hspace{.15cm}38  & \hspace{.15cm}38   & \hspace{.12cm}\underline{64}               & \hspace{.25cm}55                \\
			\textbf{4-way Junction} & \hspace{.15cm}65  & \hspace{.15cm}70   & \hspace{.12cm}\underline{71}                 & \hspace{.25cm}67                \\
			\textbf{T-junction} & \hspace{.15cm}52  & \hspace{.15cm}44   & \hspace{.12cm}41                 & \hspace{.25cm}\underline{68}                
			\\\hline
			\textbf{4 Class Avg.} & \hspace{.15cm}55  & \hspace{.15cm}55   & \hspace{.12cm}59                 & \hspace{.25cm}\underline{\textbf{62}}
			\\\hline
			\textbf{Undefined}      & \hspace{.15cm}\underline{43}  & \hspace{.15cm}39   & \hspace{.12cm}33                 & \hspace{.25cm}39    
		\end{tabular}}
	\end{table}
	
	\begin{table}[h!]
		\caption{4 Class (see Section \ref{open}) scene classification success (\%) on the Nordland dataset: state-of-the-art vs. ours.}
		\centering
		\label{table:POI_detection_N}
		\scalebox{1.2}{
			\begin{tabular}{l|l|l|l|l}
				& \hspace{.0cm} \textbf{365} \hspace{.0cm} & \textbf{1365} & \hspace{.1cm}\textbf{P2}& \textbf{EEOR} \\\hline
				\textbf{Bridges} & \hspace{.15cm}56  & \hspace{.15cm}56   & \hspace{.12cm}\underline{71}              & \hspace{.25cm}47               \\
				\textbf{Level Crossing} & \hspace{.15cm}47  & \hspace{.15cm}52   & \hspace{.12cm}56                 & \hspace{.25cm}\underline{58}              \\
				\textbf{Stations} & \hspace{.15cm}83  & \hspace{.15cm}\underline{90}   & \hspace{.12cm}89                 & \hspace{.25cm}78                \\
				\textbf{Tunnels} & \hspace{.15cm}67  & \hspace{.15cm}60   & \hspace{.12cm}59                 & \hspace{.25cm}\underline{69}               
				\\\hline
				\textbf{4 Class Avg.} & \hspace{.15cm}63  & \hspace{.15cm}65   & \hspace{.12cm}\underline{\textbf{71}}                 & \hspace{.25cm}61
				\\\hline
				\textbf{Undefined}      & \hspace{.15cm}65  & \hspace{.15cm}\underline{66}   & \hspace{.12cm}51                 & \hspace{.25cm}57    
			\end{tabular}}
		\end{table} 
\subsubsection{Does This Approach Work for Open Set Classification?}
\label{open}
The vast majority of scene recognition approaches do not consider open set classification, but specifically for this task the model should be able to identify `undefined' images that do not belong to any scene class. Open set classification for scene recognition and VPR remains a very challenging area, but for completeness the results were included as an aside in Table \ref{table:POI_detection_STL} and Table \ref{table:POI_detection_N}. The `undefined' class was trained as a scene class of its own. When training P2 a 'undefined' class was not available, which may help explain that network's poor open set classification accuracy. There is scope for using triplet learning to address this problem, as in \cite{Miller}.

\subsection{Visual Mapping: Which Approach Maps the Most Target Scene Classes?}
\label{ContextVisualMapCreation}
For this experiment the St. Lucia dataset was left untouched and the Nordland dataset was mapped at intervals of 100m to reduce computational load, resulting in a dataset of 6,546 images. Tunnel interiors were also removed from Nordland. Approximately 10\% of both datasets consisted of scene classes. One of the three traversals of each dataset was randomly selected for mapping and the visual map size was fixed to 50\% of its total frames. 

\begin{figure}[h!]
	\centering
	\includegraphics[scale=0.55]{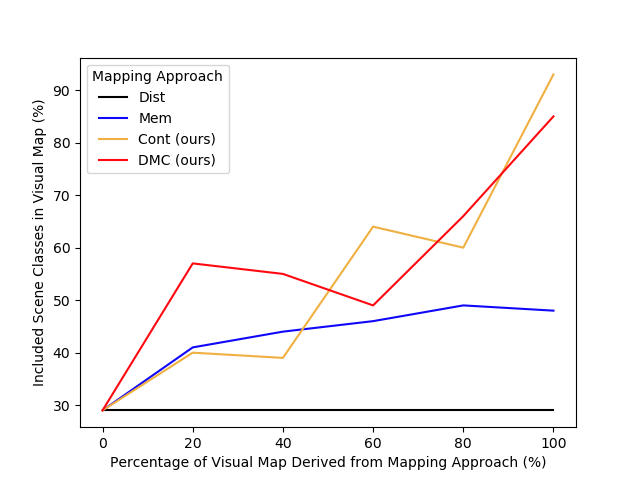}
	\caption{Percentage of available scene classes included in visual map using four different mapping techniques on the St. Lucia dataset.}
	\label{fig:map_POI_coverage_STL}
\end{figure}

\begin{figure}[h!]
	\includegraphics[scale=0.53]{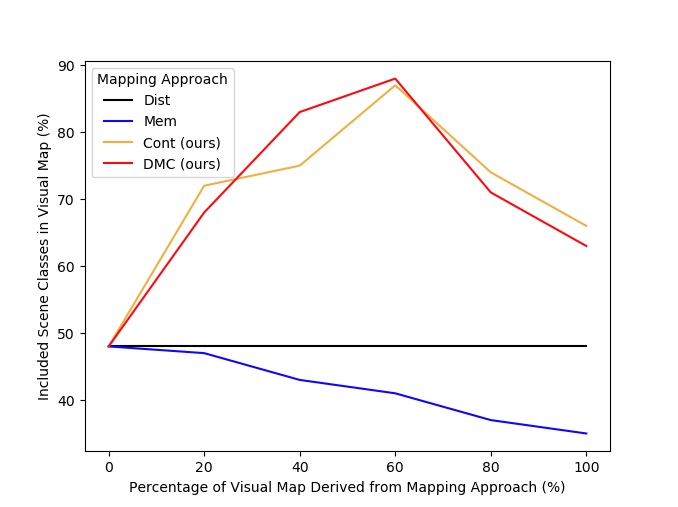}
	\caption{Percentage of available scene classes included in visual map using four different map mapping techniques on the Nordland dataset}
	\label{fig:map_POI_coverage_N}
\end{figure}

Only targetting scene classes when creating a visual map is impractical so distance (Dist) interval mapping alone was used to create a visual map and the memorability (Mem) and context (Cont) were used to source progressively larger amounts of the visual map while distance mapping was used to fill the rest. The purpose of this was to see what percentage of the scene classes could be identified and how much of the map needed to be dedicated to this task. This paper's DMC algorithm was also used for comparison with its parameters set to contribute identical amounts of frames from its combination of memorability and context with the rest sourced from its distance interval mapping, as seen in Figures \ref{fig:map_POI_coverage_STL} \& \ref{fig:map_POI_coverage_N}. 

For the St. Lucia dataset EEOR was used to provide context and 'P2' for Nordland. For approaches using context 25\% of the scene class frames and 10\% of the 'undefined' frames were randomly selected for reference scene class images. These frames were removed from the map and not used for calculating how many of the available scene classes were included in the map. The missing frames were compensated for in the map size. 

\subsubsection{Can Memorability Be Used to Select Target Scene Classes?}
For the St. Lucia dataset creating a visual map with 80\% of its frames sourced according to their memorability includes 19\% more of the scene classes, whereas for Nordland it reduces their inclusion by 13\%, compared to standard distance interval mapping. This shows that memorability alone cannot be used to reliably include more scene class frames in a visual map.  

\subsubsection{Can Context Be Used to Select Target Scene Classes?}
Contextual mapping from scene retrieval increases inclusion of scene classes significantly compared to the distance interval mapping baseline. It's maximum performance on the St. Lucia dataset is a selection of 45\% more of the scene classes when the entire visual map is created according to contextual mapping. For Nordland when 60\% of the visual map is created using contextual mapping 36\% more of the scene classes are included in the map. This amount decreases as more frames are selected according to their context. This is due to false positives being selected and scene class images which were previously included from the distance mapping being discarded in favour of better contextual matches.

Contextual classifications were calculated using a confidence threshold so we also observe that EEOR is less well calibrated than P2 is for Nordland, but that it is generally more accurate because lowering the confidence threshold allows more true positives to be classified. For Nordland P2 is better calibrated because more of its confident matches are true positives, but as the confidence threshold is lowered more false positive classifications are made. 

\begin{table*}[b!]
	\caption{Percentage difference in localization accuracy of unidentified (\textcolor{blue}{blue}) and scene class (\textcolor{red}{red}) with mapping techniques compared to a distance interval mapping baseline on two St. Lucia test traversals, as described in Section \ref{ContextVisualMap}.}
	\centering
	\label{table:table1}
	\scalebox{1.24}{
	\begin{tabular}{l|l|l|l|l|l|l|l|l|l}
		 & \multicolumn{3}{l|}{AMOS-Net}                                                    & \multicolumn{3}{l|}{Hybrid-Net}                                                  & \multicolumn{3}{l}{NetVLAD}                                                    \\
		(\%) of Vis. Map$\,\to\,$ & \multicolumn{1}{l|}{20}    & \multicolumn{1}{l|}{60}   & \multicolumn{1}{l|}{100}   & \multicolumn{1}{l|}{20}    & \multicolumn{1}{l|}{60}   & \multicolumn{1}{l|}{100}   & \multicolumn{1}{l|}{20}    & \multicolumn{1}{l|}{60}   & \multicolumn{1}{l}{100}   \\\hline
		Memorability        & \textcolor{blue}{-15}  \textcolor{red}{24} & \textcolor{blue}{-45}  \textcolor{red}{16} & \textcolor{blue}{-66}  \textcolor{red}{-8.0} & \textcolor{blue}{-14}  \textcolor{red}{20} & \textcolor{blue}{-37}  \textcolor{red}{19} & \textcolor{blue}{-67}  \textcolor{red}{-26} & \textcolor{blue}{\hspace{.08cm}0.0}  \textcolor{red}{\hspace{.09cm}5.3} & \textcolor{blue}{\hspace{.03cm}-0.7}  \textcolor{red}{4.6} & \textcolor{blue}{-7.1}  \textcolor{red}{-6.9} \\
		Context (\textit{ours})    & \textcolor{blue}{-28} \textcolor{red}{31} & \textcolor{blue}{-45} \textcolor{red}{55} & \textcolor{blue}{-77} \textcolor{red}{\hspace{.12cm}65} & \textcolor{blue}{-25}  \textcolor{red}{25} & \textcolor{blue}{-51}  \textcolor{red}{46} & \textcolor{blue}{-76}  \textcolor{red}{\hspace{.1cm}61} & \textcolor{blue}{\hspace{.08cm}0.0}  \textcolor{red}{-6.9} & \textcolor{blue}{-1.4}  \textcolor{red}{\hspace{.06cm}23} & \textcolor{blue}{-3.6}  \textcolor{red}{\hspace{.1cm}38} \\
		DMC (\textit{ours})  & \textcolor{blue}{-15} \textcolor{red}{41} & \textcolor{blue}{-32} \textcolor{red}{44} & \textcolor{blue}{-91} \textcolor{red}{\hspace{.12cm}63} & \textcolor{blue}{-14}  \textcolor{red}{36} & \textcolor{blue}{-27}  \textcolor{red}{36} & \textcolor{blue}{-88}  \textcolor{red}{\hspace{.08cm}58} & \textcolor{blue}{-0.7}  \textcolor{red}{\hspace{.08cm}3.1} & \textcolor{blue}{-2.9}  \textcolor{red}{\hspace{.06cm}21} & \textcolor{blue}{\textbf{-3.6}}  \textcolor{red}{\textbf{\hspace{.1cm}41}}
	\end{tabular}} 
\end{table*}
	
\begin{table*}[b!]
\caption{Percentage difference in localization accuracy of unidentified (\textcolor{blue}{blue}) and scene class (\textcolor{red}{red}) with mapping techniques compared to a distance interval mapping baseline on two Nordland test traversals, as described in Section \ref{ContextVisualMap}.}
\centering
\label{table:table2}
\scalebox{1.12}{
	\begin{tabular}{l|l|l|l|l|l|l|l|l|l}
		  & \multicolumn{3}{l|}{AMOS-Net}                                                    & \multicolumn{3}{l|}{Hybrid-Net}                                                  & \multicolumn{3}{l}{NetVLAD}                                                    \\
		(\%) of Vis. Map$\,\to\,$ & \multicolumn{1}{l|}{20}    & \multicolumn{1}{l|}{60}   & \multicolumn{1}{l|}{100}   & \multicolumn{1}{l|}{20}    & \multicolumn{1}{l|}{60}   & \multicolumn{1}{l|}{100}   & \multicolumn{1}{l|}{20}    & \multicolumn{1}{l|}{60}   & \multicolumn{1}{l}{100}   \\\hline
		Memorability        & \textcolor{blue}{-0.8}  \textcolor{red}{-8.6} & \textcolor{blue}{-12}  \textcolor{red}{\hspace{.06cm}-15} & \textcolor{blue}{\hspace{.15cm}n/a}  \textcolor{red}{\hspace{.12cm}n/a} & \textcolor{blue}{-5.5}  \textcolor{red}{-4.1} & \textcolor{blue}{-21}  \textcolor{red}{-12} & \textcolor{blue}{\hspace{.1cm}n/a}  \textcolor{red}{\hspace{.16cm}n/a} & \textcolor{blue}{\hspace{.1cm}0.3}  \textcolor{red}{\hspace{.08cm}0.0} & \textcolor{blue}{-5.2}  \textcolor{red}{-9.6} & \textcolor{blue}{\hspace{.1cm}n/a}  \textcolor{red}{\hspace{.07cm}n/a} \\
		Context (\textit{ours})      & \textcolor{blue}{-14} \textcolor{red}{\hspace{.08cm}-18} & \textcolor{blue}{-34} \textcolor{red}{-4.6} & \textcolor{blue}{-491} \textcolor{red}{\hspace{.03cm}-14} & \textcolor{blue}{-16}  \textcolor{red}{\hspace{.06cm}-7.7} & \textcolor{blue}{-34}  \textcolor{red}{-0.1} & \textcolor{blue}{-507}  \textcolor{red}{-3.9} & \textcolor{blue}{-3.6}  \textcolor{red}{-3.7} & \textcolor{blue}{-8.0}  \textcolor{red}{\hspace{.08cm}3.8} & \textcolor{blue}{-274}  \textcolor{red}{4.3} \\
		DMC (\textit{ours})  & \textcolor{blue}{-9.5} \textcolor{red}{-8.6} & \textcolor{blue}{-20} \textcolor{red}{-4.5} & \textcolor{blue}{-430} \textcolor{red}{-9.4} & \textcolor{blue}{-12}  \textcolor{red}{\hspace{.03cm}-3.0} & \textcolor{blue}{-22}  \textcolor{red}{\hspace{.05cm}3.4} & \textcolor{blue}{-445}  \textcolor{red}{-4.3} & \textcolor{blue}{-2.2}  \textcolor{red}{-7.4} & \textcolor{blue}{\textbf{-4.4}}  \textcolor{red}{\textbf{\hspace{.08cm}4.9}} & \textcolor{blue}{-238}  \textcolor{red}{2.6}
	\end{tabular}} 
\end{table*}
		
\subsection{Does Memorability Decrease DMC's Performance?}
Memorability is included in our DMC algorithm to increase localization accuracy, not to increase the amount of scene classes included in the map. For St. Lucia memorability correlates with the chosen scene classes so it includes 25\% more scene class images for a 20\% contribution to the visual map, but as memorability becomes less relevant it slightly decreases performance by 11\%. For Nordland on average it only results in a decrease of 1\% inclusion of scene classes. This shows that with the correct biasing constant memorability can affect performance negatively very little. Some variation in performance between DMC and contextual mapping should be expected as they each use randomly selected reference frames.

\subsection{Visual Mapping: Do Better Maps Mean Better Localization Accuracy?}
\label{ContextVisualMap}
The maps created in Section \ref{ContextVisualMapCreation} were then used to localize each dataset's two remaining routes using NetVLAD, AMOS-Net and Hybrid-Net descriptors. Localization was done using a windowed search of approximately 100 nearby frames, otherwise false positives overwhelmed the results. The results in Tables \ref{table:table1} \& \ref{table:table2} were reported as a percentage change in localization accuracy with respect to localization using the baseline visual map created with distance interval mapping. 

\begin{figure}[h!]
	\centering
	\includegraphics[scale=0.12]{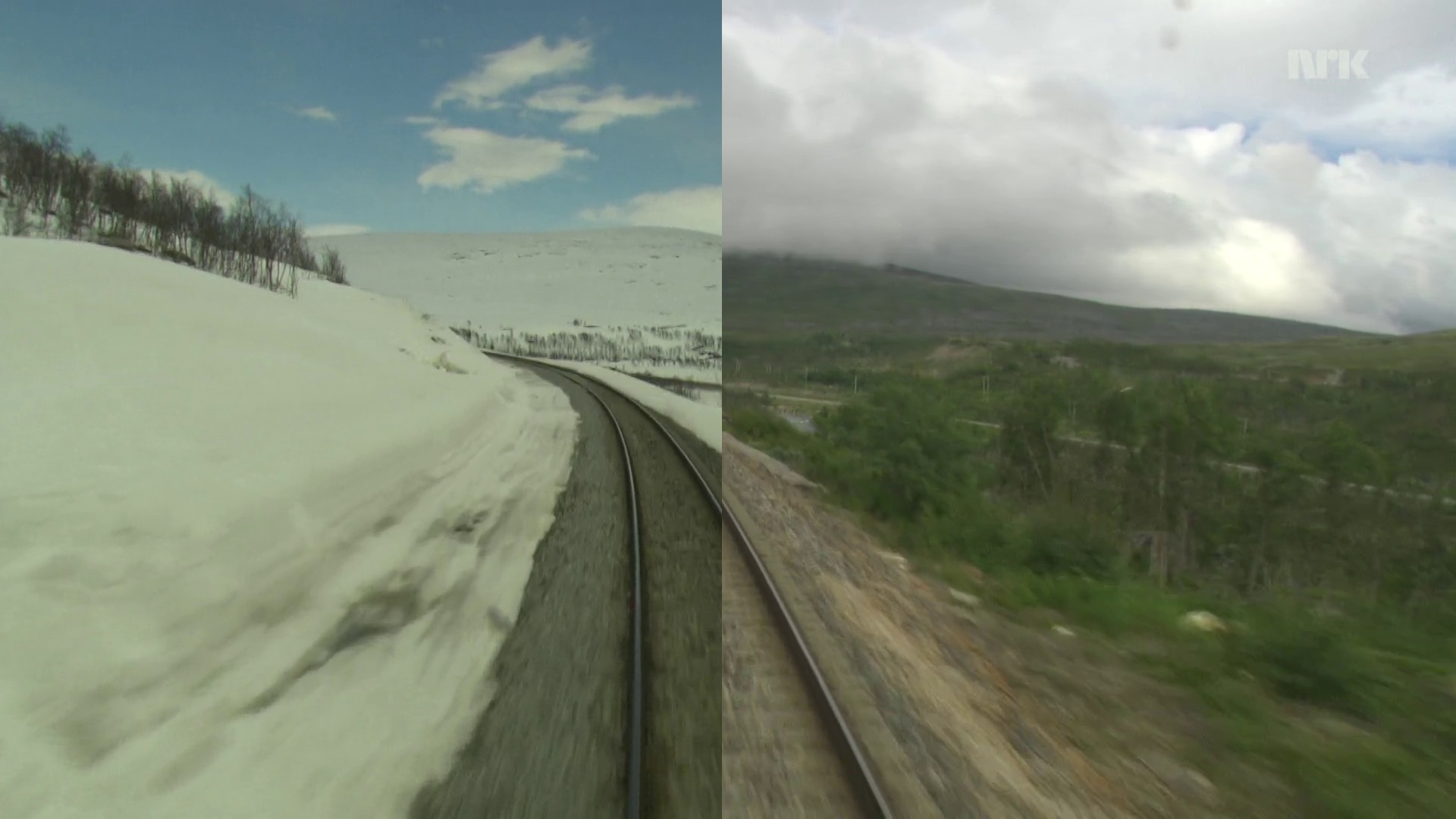}
	\caption{Example of challenging localization conditions in the Nordland dataset, this is the same place across two of our example traversals.}
	\label{fig:Nordland_challenge}
\end{figure}

\subsubsection{Does Contextual Mapping Work For Actual Localisation?}
Contextual mapping works consistently well for the St. Lucia dataset: scene class localization accuracy is improved on all but one of the visual maps using context with the highest increase in scene class localization of 65\% using AMOSNet descriptors and contextual sampling to create the entire visual map. 

DMC achieves a more modest increase in scene class localization accuracy of 4.9\% on the Nordland dataset when it used to create 60\% of the visual map. Poor map quality is not the cause for this because Figure \ref{fig:map_POI_coverage_N} shows the large amount of available scene classes in the map, instead the challenging nature of this dataset is the cause. Visual differences between test data and map data are greater in the Nordland dataset than St. Lucia, for an example see figure \ref{fig:Nordland_challenge}. Despite the windowed search each descriptor is still localized against an area over approximately 10km. We show in Table \ref{table:table1} that DMC is agnostic to the descriptor used to improve scene class localization performance, but that does not mean it is capable of compensating for poor descriptor performance.

\subsubsection{Does Memorability Improve DMC's Localisation Performance?}
Increasing scene selection in visual maps corresponds with a decrease in undefined scene class localization accuracy, which is inevitable given the fixed map size and the redistribution of map frames around the scene classes. Regardless of memorability's ability to classify scene classes, once contextual mapping has identified them it's inclusion in DMC means it is able to offset this effect to some degree. We use it to remove misleading images and improve scene class localization accuracy by a mean of 3\% and undefined frame localization accuracy of 10\% across both datasets compared to just using contextual mapping. 

Although the highest increase in scene localization is 65\% using contextual mapping to create 100\% of the visual map the overall best result uses NetVLAD features and DMC to create 100\ of the visual  that increases scene class localization accuracy by 41\% with a decrease of only 3.6\% accuracy of undefined scene classes.

\section{CONCLUSIONS}
This paper formulates a 'scene retrieval' task for classification of scenes defined at test time. We addresses this with a neural network model taught using triplet loss that improves classification accuracy by up to 7\%, in comparison to state-of-the-art networks taught on a similar task. Scene retrieval is then used as context for visual map sampling. Context is combined with memorability and distance sampling for a visual mapping algorithm DMC that increases scene class inclusion in visual maps by up to 64\% in comparison to just using distance interval mapping. The addition of memorability is shown to increase scene class localization accuracy by a mean of 3\% and undefined frame localization accuracy by 10\% across both datasets compared to just using contextual mapping. 

Future work aims to increase the generalisation of scene retrieval to scene classes across different datasets and address the problem of open set classification for visual navigation. 


\addtolength{\textheight}{-12cm}   







\bibliographystyle{IEEEtran}
\bibliography{IEEEabrv,ICRA_2020_trimmed}

\end{document}